\documentclass[9pt,journal,twoside,web]{ieeecolor}
\usepackage{generic}
\usepackage{cite}
\usepackage{amsmath,amssymb,amsfonts}
\usepackage{algorithmic}
\usepackage{textcomp}
\usepackage{times}
\usepackage{latexsym}
\usepackage{makecell}
\usepackage{here}
\usepackage{stmaryrd}
\usepackage{multirow}
\usepackage{subfig}
\usepackage{subfig}
\usepackage{dblfloatfix}
\usepackage{changepage}
\usepackage{threeparttable}
\usepackage{tabu}
\usepackage{longtable}
\usepackage{float}
\usepackage{colortbl}
\usepackage[dvipsnames,table]{xcolor}
\usepackage{booktabs}
\usepackage{tabularx}
\usepackage{longtable}

\usepackage{array}
\usepackage[utf8]{inputenc} 
\usepackage[T1]{fontenc}    
\usepackage{hyperref}       
\usepackage{url}            
\usepackage{booktabs}       
\usepackage{nicefrac}       
\usepackage{microtype}      
\usepackage{lipsum}
\usepackage{graphicx}
\graphicspath{ {./images/} }
\def\BibTeX{{\rm B\kern-.05em{\sc i\kern-.025em b}\kern-.08em
    T\kern-.1667em\lower.7ex\hbox{E}\kern-.125emX}}
\begin{document}
\title{Explainable Identification of Dementia from Transcripts using Transformer Networks}
\author{Loukas Ilias and Dimitris Askounis
\thanks{}
\thanks{Loukas Ilias and Dimitris Askounis are with the Decision Support Systems Laboratory, School of Electrical and Computer Engineering, National Technical University of Athens, 15780 Athens, Greece (e-mail: \{lilias,askous\}@epu.ntua.gr). }
}

\maketitle

\begin{abstract}
Alzheimer's disease (AD) is the main cause of dementia which is accompanied by loss of memory and may lead to severe consequences in peoples' everyday life if not diagnosed on time. Very few works have exploited transformer-based networks and despite the high accuracy achieved, little work has been done in terms of model interpretability. In addition, although Mini-Mental State Exam (MMSE) scores are inextricably linked with the identification of dementia, research works face the task of dementia identification and the task of the prediction of MMSE scores as two separate tasks.  In order to address these limitations, we employ several transformer-based models, with BERT achieving the highest accuracy accounting for 87.50\%. Concurrently, we propose an interpretable method to detect AD patients based on siamese networks reaching accuracy up to 83.75\%. Next, we introduce two multi-task learning models, where the main task refers to the identification of dementia (binary classification), while the auxiliary one corresponds to the identification of the severity of dementia (multiclass classification). Our model obtains accuracy equal to 86.25\% on the detection of AD patients in the multi-task learning setting. Finally, we present some new methods to identify the linguistic patterns used by AD patients and non-AD ones, including text statistics, vocabulary uniqueness, word usage, correlations via a detailed linguistic analysis, and explainability techniques (LIME). Findings indicate significant differences in language between AD and non-AD patients.
\end{abstract}


\section{Introduction}
\label{sec:introduction}
Alzheimer's disease (AD) constitutes a neurodegenerative disease characterized by a progressive cognitive decline and is the leading cause of dementia. Signs of dementia include amongst others: problems with short-term memory, keeping track of a purse or wallet, paying bills, planning and preparing meals, remembering appointments, or travelling out of the neighborhood \cite{WinNTT}. Because of the fact that Alzheimer's dementia gets worse over time, it is important to be diagnosed early. For this reason, several research works have been introduced targeting at diagnosing dementia, which use imaging techniques \cite{ZHANG2019185}, CSF biomarkers \cite{hassan2017machine,DAVATZIKOS20112322.e19}, or EEG signals \cite{IERACITANO2020176}. Due to the fact that dementia affects speech to a high degree, recently the research has moved towards dementia identification from spontaneous speech, where several shared tasks \cite{bib:LuzHaiderEtAl20ADReSS,luz21_interspeech} have been developed in order to distinguish AD from non-AD patients.

Several research works have been conducted with regard to the identification of AD patients using speech and transcripts. The majority of them have employed feature extraction techniques \cite{weiner2018selecting,CALZA2021101113,10.3389/fnagi.2019.00205,10.3389/fcomp.2021.640669,khodabakhsh2014natural}, in order to train traditional Machine Learning (ML) algorithms, such as Logistic Regression, k-NN, Random Forest, etc. However, feature extraction constitutes a time-consuming procedure achieving poor classification results and often demands some level of domain expertise. Recently, researchers introduce deep learning architectures \cite{chen2019attention,di-palo-parde-2019-enriching}, such as CNNs and BiLSTMs, so as to improve the classification results. Despite the success of transformer-based models in several domains, their potential has not been investigated to a high degree in the task of dementia identification from transcripts, where research works \cite{syed2021automated} having proposed them, use their outputs as features to train shallow machine learning algorithms. Concurrently, all research works except one \cite{10.3389/fcomp.2021.624683}, train machine learning models, in order to distinguish AD patients from non-AD patients, without taking into account the severity of dementia via Mini-Mental State Exam (MMSE) scores. Motivated by this limitation, we propose two multi-task learning models minimizing the loss of both dementia identification and its severity. 

At the same time, to the best of our knowledge, the research works that have proposed deep learning models based on transformer networks have focused their interest only on improving the classification results obtained by CNNs, BiLSTMs etc. instead of exploring possible explainability techniques. Specifically, due to the fact that deep learning models are considered black boxes, it is important to propose ways of making them interpretable, since it is imperative for a clinician to be informed why the specific deep neural network classified a person as AD patient or not. To the best of our knowledge, only one work \cite{karlekar-etal-2018-detecting} has experimented with interpreting its proposed deep learning model (CNN-LSTM model) in the field of dementia detection using transcripts. In order to tackle this limitation, our contribution is twofold. First, we propose an interpretable neural network architecture. Next, we extend prior work and employ LIME \cite{ribeiro2016should}, a model agnostic framework for interpretability, aiming to explain the predictions made by our best performing model. Concurrently, we propose an in-depth analysis of the language patterns used between AD and non-AD patients aiming to shed more light on the main differences observed in the vocabulary that may distinguish people suffering from dementia from healthy people.  

Our main contributions can be summarized as follows:
\begin{itemize}
    \item We employ several transformer-based models, pretrained in biomedical and general corpora, and compare their performances.
    \item We propose an interpretable method based on the siamese neural networks along with a co-attention mechanism, so as to detect AD patients.
    \item We introduce two models in a multi-task learning framework, where the one task is the identification of dementia and the second one is the detection of MMSE score (severity of dementia). We model the MMSE detection task as a multiclass classification task instead of a regression task.
    \item We perform a thorough linguistic analysis regarding the differences in language between control and dementia groups.
    \item We employ LIME, in order to explain the predictions of our best performing model.
\end{itemize} 

\section{Related Work}
\subsection{Feature-based} 

The authors in \cite{10.3389/fpsyg.2020.624137,10.3389/fcomp.2021.624659} introduced approaches based on multimodal data (both linguistic and acoustic features) to detect AD patients (binary classification task) and predict MMSE score (regression task). More specifically, the authors in \cite{10.3389/fpsyg.2020.624137} exploited dimensionality reduction techniques followed by machine learning classifiers and stated that Logistic Regression (LR) with language features was their best performing model in terms of classifying AD and non-AD patients. With regards to estimating the MMSE score, they claimed that a Random Forest classifier with language features achieves the lowest RMSE and $R^2$ scores. The combination of linguistic and acoustic features did not perform well on both tasks. In \cite{10.3389/fcomp.2021.624659}, the authors trained both shallow and deep learning models (LSTM and CNN) on a feature set consisting of acoustic features (i-vectors, x-vectors) and text features (word vectors, BERT embeddings, LIWC features, and CLAN features) to detect AD patients. They found that the top-performing classification models were the Support Vector Machine (SVM) and Random Forest classifiers trained on BERT embeddings, which both achieved an accuracy of 85.4\% on the test set. Regarding the regression task, they claimed that the gradient boosting regression model using BERT embeddings outperformed all the other introduced architectures. Authors in \cite{syed2021automated} trained shallow machine learning algorithms (Logistic Regression and Support Vector Machine for detecting AD patients, and Support Vector Machines based regression and Partial Least Squares Regressor for predicting the MMSE scores) using embeddings extracted by transformer-based models, namely BERT, RoBERTa,  DistilBERT, DistilRoBERTa, and BioMed-RoBERTa-base. A similar approach was conducted by \cite{syed20_interspeech}, where the authors extracted embeddings for each word of the transcript using transformer-based networks, exploited four types of pooling functions for generating a transcript-level representation, and trained a Logistic Regression classifier. Research work \cite{pompili20_interspeech} merged acoustic (x-vectors) and linguistic features and trained a Support Vector Machine Classifier. In terms of the language features, (i) a Global Maximum pooling, (ii) a bidirectional LSTM-RNNs provided with an attention module, and (iii) the second model augmented with part-of-speech (POS) embeddings were trained on the top of a pretrained BERT model. Nasreen et al. \cite{10.3389/fcomp.2021.640669} extracted two feature sets, namely disfluency and interactional features, and performed an in-depth statistical analysis in an attempt to investigate the differences between AD and non-AD subjects in terms of these features. Findings show that these two groups of people present significant differences. Then, they exploited shallow machine learning algorithms using the aforementioned feature sets to distinguish AD from non-AD patients and obtained an accuracy of 0.90 when providing both feature sets as input to the SVM classifier.

\subsection{Deep Learning} 
Research works \cite{pan2019automatic,kong2019neural} employed a hierarchical attention neural network to detect AD patients. More specifically, the authors in \cite{pan2019automatic} evaluated their proposed model in both manual and automatic transcripts and found that a hierarchical neural network achieves an improvement in F1-score in comparison to other deep learning models. In \cite{kong2019neural}, the authors tried to interpret the decisions made by the proposed model by visualizing words and sentences and performing statistical analyses. However, they were not able to explain why their model pays attention to some specific words more than others. Moreover, an explainable approach was introduced by \cite{karlekar-etal-2018-detecting}. Specifically, after proposing three deep learning architectures based on CNNs and RNNs, the authors applied visualization techniques and showed which linguistic characteristics are indicative of dementia, i.e., short answers, repeated requests for clarification, and interjections at the start of each utterance. Authors in \cite{pan2021multi} proposed a multi-task learning framework (Sinc-CLA), so as to predict age and MMSE scores (both considered as regression tasks) and used only speech as input for their proposed network. Concurrently, they introduced shallow networks with input i-vectors and x-vectors both in single and multi-task learning frameworks. They claimed that using x-vectors in a multi-task learning framework yields the best results in terms of the estimation of both age and MMSE scores. Ref. \cite{balagopalan20_interspeech} introduced both feature-based and transformer-based methods. Regarding transformer-based models, they fine-tuned the BERT model to detect AD patients achieving better evaluation results than the ones achieved via the feature-based methods. For estimating the MMSE score they proposed only feature-based approaches.  Research work \cite{10.3389/fcomp.2021.624683} is the most similar to ours. The authors proposed transformer-based models using text, audio, and images (they converted audio to images using Mel Frequency Cepstral Coefficient). Regarding text, they employed BERT and Longformer. They claimed that models using only text data outperformed all the other proposed ones. The fusion of text and audio did not achieve better results. They introduced also a multi-task learning architecture using only text as input, in order to predict the MMSE score (regression task) and detect AD patients (binary classification task). Results showed limited improvements in classification and a negative impact in regression. We extend this research work by employing more transformer-based networks with an efficient training strategy, proposing a new interpretable method to detect AD patients based on siamese networks, introducing two models in a multi-task learning framework by regarding the MMSE prediction task as a multiclass classification task and employing explainability techniques. On the other hand, research works \cite{10.3389/fcomp.2021.624558} \& \cite{rohanian20_interspeech} introduced deep learning models including CNNs and LSTM neural networks with feed-forward highway layers respectively. In \cite{10.3389/fcomp.2021.624558} results suggested that the utterances of the interviewer boost the classification performance. A similar methodology with \cite{rohanian20_interspeech} was proposed by \cite{rohanian21_interspeech}, where the authors exploited both BERT and LSTMs with gating mechanism and showed that LSTM with gating mechanism outperforms BERT model with gating mechanism. They stated that this difference may be attributable to the fact that BERT is very large in comparison to the LSTM models. Researchers in \cite{cummins20_interspeech} introduced four approaches for detecting AD patients. Specifically, they trained a hierarchical neural network with an attention mechanism on linguistic features. Concurrently, they proposed a Siamese Neural Network and a Convolutional Neural Network using audio waveforms. Finally, they extracted features from audio segments and trained an SVM classifier. Results showed that the combination of audio features, CNNs, and hierarchical neural network achieved the best classification results.

\subsection{Related Work Review Findings}

From the aforementioned research works, it is evident that despite the negative consequences dementia has in people's everyday life, little work has been done so far towards its identification. More specifically, most researchers introduce feature extraction approaches from audio and transcripts and train ML algorithms, such as SVM, LR, etc. Because of the fact that feature extraction constitutes a time-consuming procedure and does not generalize well to new AD patients, researchers have started exploiting deep learning methods, such as CNNs and LSTMs, which obtain low performances. However, despite the fact that pretrained transformer models achieve new state-of-the-art results in several domains, including the biomedical one, their potential has been mainly used as embeddings for training shallow ML algorithms, such as SVM or LR. Concurrently, little has been done regarding the interpretability of the proposed deep learning models as well as the main differences observed in the language between AD patients and non-AD patients. 

Our work is different from the research works mentioned above, since we: \textit{(a)} propose several pretrained transformer-based models and compare their performances,  \textit{(b)} introduce the idea of siamese neural networks along with a co-attention mechanism towards the task of dementia classification,  \textit{(c)} convert the MMSE regression task into a multiclass classification one and explore if it helps dementia identification, \textit{(d)} perform a detailed linguistic analysis to find the linguistic patterns that distinguish AD patients from non-AD ones, and \textit{(e)} exploit LIME for explaining the predictions made by our best performing model.

\section{Dataset}
We use the ADReSS Challenge Dataset \cite{bib:LuzHaiderEtAl20ADReSS} for conducting our experiments. In contrast to other datasets, this dataset is matched for gender and age, so as to minimize the risk of bias in the prediction tasks. Moreover, it has been selected in such a way so as to mitigate biases often overlooked in evaluations of AD detection methods, including repeated occurrences of speech from the same participant (common in longitudinal datasets) and variations in audio quality. It consists of speech recordings along with their associative transcripts and includes 78 non-AD and 78 AD subjects. In addition, the dataset includes the MMSE scores for each subject except one. We report the mean and standard deviation of the MMSE scores for the two main groups, i.e., AD patients and non-AD ones, in Table \ref{descriptive_purposes}. Each participant (PAR) has been assigned by the interviewer (INV) to describe the Cookie Theft picture from the Boston Diagnostic Aphasia Exam \cite{10.1001/archneur.1994.00540180063015}. Due to the fact that the transcripts are annotated using the CHAT coding system \cite{macwhinney2014childes}, we use the python library PyLangAcq \cite{lee-et-al-pylangacq:2016} for having access to the dataset. We use data (utterances) only from PAR and conduct our experiments at the transcript-level. The ADReSS Challenge dataset has been divided into a train and a test set. The train set consists of 54 AD patients and 54 non-AD ones, while the test set consists of 24 AD patients and 24 non-AD ones.

\begin{table}[hbt]
\caption{Mean and standard deviation of the MMSE scores for the two main groups (AD and non-AD patients).}
\centering
\begin{tabular}{lcc}
\toprule
\multicolumn{1}{l}{}&\multicolumn{2}{c}{\textbf{MMSE}}\\
\cline{2-3}
& \textbf{mean} & \textbf{standard deviation} \\ \hline
\textbf{AD} & 17.79 & 5.48 \\ \hline
\textbf{non-AD} & 29.01 & 1.17 \\
\bottomrule
\end{tabular}
\label{descriptive_purposes}
\end{table}

\section{Problem Statement}
In this section, the problem statement used in this paper is presented. More specifically, it can be divided into two problems, namely the Single-Task Learning (STL) Problem and the Multi-Task Learning (MTL) Problem, which are presented in detail in Sections \ref{stl_problem} and \ref{mtl_problem} respectively.
\subsection{Single-Task Learning Problem} \label{stl_problem}

Let a dataset \( \mathcal{S}_{n \times 2}=\begin{bmatrix} s_1,label_1\\s_2,label_2\\ \vdots \\s_n,label_n \end{bmatrix} \) consist of a set of transcriptions belonging to the dementia group, \( d \subset \mathcal{S} \), and a set of transcriptions belonging to the control group, \( c \subset \mathcal{S} \). Furthermore, \(label_i \in \{0,1\}, 1\leq i \leq n\), where \textit{0} denotes that \( s_i \in c \), while \textit{1} denotes that \( s_i \in d \). The task is to identify if a transcription \( s_i \in \mathcal{S} \), belongs to a person suffering from dementia, i.e., \( s_i \in d \), or not, i.e., \( s_i \in c \).

\subsection{Multi-Task Learning Problem} \label{mtl_problem}

Let a dataset \( \mathcal{S}_{n \times 3}=\begin{bmatrix} s_1,label_1,mmse_1\\s_2,label_2,mmse_2\\ \vdots \\s_n,label_n,mmse_n \end{bmatrix} \) consist of a set of transcriptions belonging to the dementia group, \( d \subset \mathcal{S} \), and a set of transcriptions belonging to the control group, \( c \subset \mathcal{S} \). Furthermore, \(label_i \in \{0,1\}, 1\leq i \leq n\), where \textit{0} denotes that \( s_i \in c \), while \textit{1} denotes that \( s_i \in d \). Moreover, \(mmse_i\) indicates the MMSE scores. The tasks here are to identify \textbf{\textit{(i)}} if a transcription \( s_i \in \mathcal{S} \), belongs to a person suffering from dementia, i.e., \( s_i \in d \), or not, i.e., \( s_i \in c \), as well as \textbf{\textit{(ii)}} to identify the MMSE scores of each person.  

\section{Predictive Models}
In this section, we describe the models used for detecting AD patients. Specifically, Section \ref{stlmodels} refers to the models employed in the single-task learning setting, whereas in Section \ref{mtl_models} we refer to the models used for jointly learning to identify AD patients and detect the severity of dementia.

\subsection{Single-Task Learning} \label{stlmodels}

\subsubsection{Transformer-based models} \label{transformer-based-models}
We exploit the following transformer-based networks in our experiments: \textbf{BERT} \cite{devlin-etal-2019-bert}, \textbf{BioBERT} \cite{lee2020biobert}, \textbf{BioClinicalBERT} \cite{alsentzer-etal-2019-publicly}, \textbf{ConvBERT} \cite{NEURIPS2020_96da2f59}, \textbf{RoBERTa}  \cite{liu2019roberta}, \textbf{ALBERT} \cite{lan2019albert}, and \textbf{XLNet} \cite{NEURIPS2019_dc6a7e65}.

Regarding our experiments, we pass each transcription through each pretrained model mentioned above. The output of each model is passed through a Global Average Pooling layer followed by two dense layers. The first dense layer consists of 128 units with a ReLU activation function and the second one has one unit with a sigmoid activation function to give the final output.

\subsubsection{Transformer-based models with Co-Attention Mechanism}

In this section, we present an interpretable method to differentiate AD from non-AD patients. First, we split each transcription \textit{s} in the dataset into two statements of equal length ($s_1$ \& $s_2$). In this way, we have to categorize a pair of statements ($s_1$ \& $s_2$) into dementia or control group. To do this, we pass $s_1$ and $s_2$ through the transformer-based models mentioned in Section \ref{transformer-based-models}, i.e., BERT, BioBERT, BioClinicalBERT, ConvBERT, RoBERTa, ALBERT, and XLNet. These models can be considered as siamese in our experiments, since we make them share the same weights. Then, we implement a co-attention mechanism introduced by \cite{NIPS2016_9dcb88e0} and adopted in several studies, including \cite{shu2019defend,lu-li-2020-gcan}, over the two embeddings of the two statements (outputs of the transformer-based models), in order to render the entire architecture interpretable.

Formally, let $x_1 ^1, x_2 ^1, x_3 ^1, ..., x_N ^1$ and $x_1 ^2, x_2 ^2, x_3 ^2, ..., x_T ^2$ be the tokens of $s_1$ and $s_2$ respectively. These tokens are passed to the transformer-based models as described via the equations below:
\begin{equation}
    C = model \left(x_1 ^1, x_2 ^1, x_3 ^1, ..., x_N ^1 \right), C \in \mathbb{R}^{d \times N}
\end{equation}
\begin{equation}
    S = model \left(x_1 ^2, x_2 ^2, x_3 ^2, ..., x_T ^2 \right), S \in \mathbb{R}^{d \times T}
\end{equation}
, where \textit{model} is one of the following: BERT, BioBERT, BioClinicalBERT, ConvBERT, RoBERTa, ALBERT, and XLNet. We have omitted the first dimension, which corresponds to the batch size. Following the methodology proposed by \cite{NIPS2016_9dcb88e0}, given the output of the model receiving the tokens of $s_1$ ($\textbf{C} \in \mathbb{R}^{d \times N}$) and the output of the model receiving the tokens of $s_2$ ($\textbf{S} \in \mathbb{R}^{d \times T}$), where  $d$ denotes the hidden size of the model, the affinity matrix $F \in \mathbb{R}^{N \times T}$ is calculated using the equation $F = \tanh \left(C^T W_l S \right)$, where $W_l \in \mathbb{R}^{d \times d}$ is a matrix of learnable parameters. Next, this affinity matrix is considered as a feature and we learn to predict the attention maps for both statements via the following,
$H^s = \tanh \left(W_s S + \left(W_c C \right)F \right)$ and $H^c = \tanh \left(W_c C + \left(W_s S \right)F^T \right)$, where $W_s, W_c \in \mathbb{R}^{k \times d}$ are matrices of learnable parameters. The attention probabilities for each word in both statements are calculated through the softmax function as follows,
$a^s = softmax \left(w_{hs} ^T H^s \right)$, $a^c = softmax \left(w_{hc} ^T H^c \right)$, where $a_s \in \mathbb{R}^{1 \times T}$ and $a_c \in \mathbb{R}^{1 \times N}$. $W_{hs}, W_{hc} \in \mathbb{R}^{k \times 1}$ are the weight parameters. Based on the above attention weights, the attention vectors for each statement are obtained by calculating the weighted sum of the features from each statement. Formally:
\begin{equation}
    \hat{s} = \sum_{i=1}^{N} a_i ^s s^i, \hat{c}=\sum_{j=1}^{T} a_j ^c c^j
\end{equation}
,where $\hat{s} \in \mathbb{R}^{1 \times d}$ and $\hat{c} \in \mathbb{R}^{1 \times d}$. Finally, these two vectors are concatenated, i.e., 
$p=\left[\hat{s},\hat{c} \right]$, where $p \in \mathbb{R}^{1 \times 2d}$ and we pass the vector \textit{p} to a dense layer with 128 units and a ReLU activation function followed by a dense layer consisting of one unit with a sigmoid activation function.

\subsection{Multi-Task Learning} \label{mtl_models}

In this section we propose two architectures based on multi-task learning \cite{caruana1997multitask} and adopt the methodology followed by \cite{rajamanickam-etal-2020-joint} \& \cite{jin-aletras-2021-modeling}. To be more precise, we employ a multi-task learning framework consisting of a primary and an auxiliary task. The identification of dementia constitutes the primary task, while the prediction of the MMSE score constitutes the auxiliary one. Our main objective is to explore whether the MMSE score helps in classifying groups into dementia or control. The introduced architectures are trained on the two tasks and updated at the same time with a joint loss:

\begin{equation}
L= \left(1 - \alpha \right)L_{dementia} + \alpha L_{MMSE}
\label{equation3}
\end{equation}
,where $L_{dementia}$ and $L_{MMSE}$ are the losses of dementia
identification and MMSE prediction tasks
respectively. $\alpha$ is a hyperparameter that controls the importance we place on each task.
We mention below the MTL architectures developed.

\paragraph{\normalsize \textbf{MTL-BERT (Multiclass)}} We pass each transcription through a BERT model (which constitutes our best performing STL model). The output of the BERT model is passed through two separate dense layers, so as to identify dementia and predict the MMSE score. For identifying dementia, we use a dense layer with 2 units and a softmax activation function and minimize the cross-entropy loss function. Regarding the estimation of the MMSE score, in contrast with previous research works, we convert the MMSE regression task into a multiclass classification task. More specifically, according to \cite{rohanian20_interspeech}, we can create 4 groups of cognitive severity: \textbf{healthy} (MMSE score $\geq$ 25), \textbf{mild dementia} (MMSE score of 21–24), \textbf{moderate dementia} (MMSE score of 10–20), and \textbf{severe dementia} (MMSE score $\leq$ 9). Thus, for classifying transcriptions into one of these 4 groups, we use a dense layer of 4 units with a softmax activation function and minimize the cross-entropy loss function.  

\paragraph{\normalsize \textbf{MTL-BERT-DE (Multiclass)}} Similarly to \cite{jin-aletras-2021-modeling}, we pass each transcription into a BERT model. The output of the BERT model is passed through two separate BERT encoders, i.e, double encoders, which are followed by dense layers so as to identify dementia and classify MMSE score into one of the four classes mentioned above.  For identifying dementia, we use a dense layer with 2 units and a softmax activation function and minimize the cross-entropy loss function. For classifying the MMSE score, we use a dense layer with 4 units and a softmax activation function and minimize the cross-entropy loss function.  

\section{Experiments}
All experiments are conducted on a single Tesla P100-PCIE-16GB GPU.

\subsection{Single-Task Learning} \label{stl_experimental_setup}

\paragraph{\normalsize \textbf{Comparison with state-of-the-art approaches}}
We compare our introduced models with the following research works, since these research works propose single-task learning models and test their proposed approaches on the ADReSS Challenge test set:
\textbf{(1)} Text \cite{syed2021automated}, \textbf{(2)} LSTM with Gating (Acoustic + Lexical + Dis) \cite{rohanian20_interspeech}, \textbf{(3)} Fusion Maj. (3-best) \cite{cummins20_interspeech}, \textbf{(4)} Logistic Regression (NLP) \cite{10.3389/fcomp.2021.624659}, \textbf{(5)} fastText, bi + trigram \cite{10.3389/fcomp.2021.624558}, \textbf{(6)} Attempt 5 \cite{syed20_interspeech}, and \textbf{(7)} Fusion of system \cite{pompili20_interspeech}.

\paragraph{\normalsize \textbf{Experimental Setup}}
Firstly, we divide the train set provided by the Challenge into a train and a validation set (65\%-35\%). Next, we train the proposed architectures five times and test them using the test set provided by the Challenge. Specifically, we freeze the weights of each pretrained model (BERT, BioBERT, BioClinicalBERT, ConvBERT, RoBERTa, ALBERT, and XLNet) and update the weights of the rest layers. In this way, these pretrained models act as fixed feature extractors. We train the proposed architectures using Adam optimizer with a learning rate of 1e-4. We apply \textit{EarlyStopping} and stop training, if the validation loss has stopped decreasing for 9 consecutive epochs. We also apply \textit{ReduceLROnPlateau}, where we reduce the learning rate by a factor of 0.2, if the validation loss has stopped decreasing for 3 consecutive epochs. When this training procedure stops, we unfreeze the weights of the pretrained models and train the entire deep learning architectures using Adam optimizer with a learning rate of 1e-5. We apply \textit{EarlyStopping} with a patience of 3 based on the validation loss. In terms of models with a co-attention mechanism, we start training the proposed architectures using Adam optimizer with a learning rate of 1e-3 and follow the same methodology. We also apply dropout after the co-attention mechanism with a rate of 0.4. For \textbf{BERT}, we have used the base-uncased model, for \textbf{BioBERT} we have used BioBERT v1.1 (+PubMed), for \textbf{ConvBERT} we have used the base model, for \textbf{RoBERTa} we have employed the base model, for \textbf{ALBERT} we have used the base-v1 model, and for \textbf{XLNet} we have used the base model. For these pretrained models, we have used the Transformers library \cite{wolf-etal-2020-transformers}.\footnote{For BioClinicalBERT we have used the model in: \small{\url{https://huggingface.co/emilyalsentzer/Bio_ClinicalBERT}}}

\paragraph{\normalsize \textbf{Evaluation Metrics}} We evaluate our results using Accuracy, Precision, Recall, F1-score, and Specificity. All these metrics have been calculated using the dementia class as the positive one.

\subsection{Multi-Task Learning}

\paragraph{\normalsize \textbf{Comparison with state-of-the-art approaches}}
For the primary task (AD Classification task), we compare our introduced models with BERT base \cite{10.3389/fcomp.2021.624683}, since this research work proposes a multi-task learning model and tests its proposed approach on the ADReSS Challenge test set.

\paragraph{\normalsize \textbf{Experimental Setup}} Firstly, we divide the train set provided by the Challenge into a train and a validation set (65\%-35\%). Next, we train the proposed architectures five times and test them using the test set provided by the Challenge. We use the Adam optimizer with a learning rate of 1e-6. We apply \textit{EarlyStopping} and stop training, if the validation loss has stopped decreasing for 8 consecutive epochs. Regarding MTL-BERT-DE (Multiclass), we freeze the weights of the shared BERT model. Moreover, because of the class imbalance of the MMSE categories, we apply balanced class weights to the loss function ($L_{MMSE}$). We set $\alpha$ of \eqref{equation3} equal to 0.1. \footnote{We used also the experimental setup of Section \ref{stl_experimental_setup}. However, lower evaluation results were achieved.}
\paragraph{\normalsize \textbf{Evaluation Metrics}} For the primary task (AD Classification task), we evaluate our results using Accuracy, Precision, Recall, F1-score, and Specificity. All these metrics have been calculated using the dementia class as the positive one.

For the auxiliary task (MMSE Classification task), we evaluate our results using the average weighted Precision, average weighted Recall, and average weighted F1-score.

\section{Results}

\subsection{Single-Task Learning Experiments}

The results of the proposed models mentioned in Section \ref{stlmodels} are reported in Table \ref{compare}. Also, Table \ref{compare} provides a comparison of our introduced models with existing research initiatives.

Regarding our proposed transformer-based models, one can easily observe that BERT obtains the highest Recall, F1-score, and Accuracy accounting for 81.66\%, 86.73\%, and 87.50\% respectively. Specifically, BERT outperforms the other introduced transformer-based models in Recall by 1.67-13.33\%, in F1-score by 2.01-10.98\%, and in Accuracy by 1.25-9.17\%. BioClinicalBERT achieves the second highest Accuracy and F1-score accounting for 86.25\% and 84.72\% respectively. Also, BioClinicalBERT obtains the highest Precision score equal to 95.03\% surpassing the other transformer-based models by 4.79-15.88\%. RoBERTa achieves comparable results to BERT and BioClinicalBERT yielding an Accuracy and F1-score of 84.16\% and 82.81\% respectively. In addition, BioBERT and ConvBERT demonstrate slight differences in Accuracy and F1-score, with BioBERT surpassing ConvBERT in both metrics. Specifically, BioBERT surpasses ConvBERT in F1-score by 0.46\% and in Accuracy by 0.84\%. Moreover, we observe that ALBERT and XLNet achieve Accuracy scores equal to 78.33\%, with ALBERT surpassing XLNet in F1-score by 2.70\%.

Regarding our proposed transformer-based models with a co-attention mechanism, they achieve lower performance than the proposed transformer-based models except for ConvBERT+Co-Attention, ALBERT+Co-Attention, and XLNet+Co-Attention. More specifically, ConvBERT+Co-Attention presents a slight surge of 0.42\% in Accuracy in comparison with ConvBERT, ALBERT+Co-Attention presents an increase in Accuracy by 1.67\% in comparison with ALBERT, and XLNet+Co-Attention demonstrates a slight increase of 0.42\% in Accuracy in comparison with XLNet. BERT+Co-Attention attains the highest F1-score and Accuracy accounting for 83.85\% and 83.75\% respectively. BERT+Co-Attention outperforms the other models in terms of F1-score by 1.42-7.43\%, and in terms of Accuracy by 1.25-5.00\%. ConvBERT+Co-Attention and BioClinicalBERT+Co-Attention demonstrate slight differences in F1-score and Accuracy, with ConvBERT+Co-Attention surpassing BioClinicalBERT+Co-Attention in F1-score by 0.44\% and in Accuracy by 0.42\%. BioBERT+Co-Attention and ALBERT+Co-Attention achieve almost equal F1-score results, with BioBERT+Co-Attention attaining a higher Accuracy score than ALBERT+Co-Attention by 1.66\%. RoBERTa+Co-Attention and XLNet+Co-Attention demonstrate low performances attaining an Accuracy of 79.16\% and 78.75\% respectively.

Overall, BERT constitutes our best performing model, since it outperforms all the other introduced models in F1-score and Accuracy. Although there are models surpassing BERT in Precision and Recall, BERT outperforms all of them in F1-score, which constitutes the weighted average of Precision and Recall. In addition, there are models that outperform BERT in Specificity. However, high specificity and low recall means that the model cannot diagnose the AD patients pretty well and consequently AD patients are misdiagnosed as non-AD ones.

In comparison with the state-of-the-art approaches, one can observe that our proposed models achieve comparable performance to or outperform previous studies. More specifically, BERT outperforms all the research works, except \cite{syed2021automated}, in terms of Accuracy by 2.08-8.33\%, in F1-score by 1.33-8.68\%, and in Recall by 2.66-14.99\%. Moreover, BERT+Co-Attention surpasses \cite{rohanian20_interspeech,10.3389/fcomp.2021.624558,pompili20_interspeech} in Accuracy by 2.50\%, 0.42\%, and 4.58\% respectively. Also, it surpasses \cite{rohanian20_interspeech,10.3389/fcomp.2021.624558,pompili20_interspeech} in Recall by 17.49\%, 5.16\%, and 9.16\% respectively. BERT+Co-Attention outperforms \cite{rohanian20_interspeech,10.3389/fcomp.2021.624558,pompili20_interspeech} in F1-score by 5.80\%, 0.85\%, and 5.59\% respectively.

\begin{table}[hbt]
\caption{Performance comparison among proposed STL models and state-of-the-art approaches on the ADReSS Challenge test set. Reported values are mean $\pm$ standard deviation. Results are averaged across five runs.}
\centering
\begin{tabular}{lccccc}
\toprule
\multicolumn{1}{l}{}&\multicolumn{5}{c}{\textbf{Evaluation metrics}}\\
\cline{2-6} 
\multicolumn{1}{l}{\textbf{Architecture}}&\textbf{Prec.}&\textbf{Rec.}&\textbf{F1-score}&\textbf{Acc.}&\textbf{Spec.}\\
\midrule
\multicolumn{6}{>{\columncolor[gray]{.8}}l}{\textbf{Comparison with state-of-the-art approaches}} \\
\small \textit{\cite{syed2021automated}} & \small- & \small87.50 & \small- & \small89.58 & \small91.67 \\ \hline
\textit{\small{\cite{rohanian20_interspeech}}} & 81.82 & 75.00 & 78.26 & 79.17 & 83.33 \\ \hline
\small \textit{\cite{cummins20_interspeech}} & \small - & \small- & \small85.40 & \small85.20 &\small -\\ \hline
\small \textit{\cite{10.3389/fcomp.2021.624659}} & \small- & \small- & \small- & \small85.00 &\small -\\ \hline 
\small \textit{\cite{10.3389/fcomp.2021.624558}} & \small86.00 & \small79.00 & \small83.00 & \small83.33 & \small88.00\\ \hline
\small \textit{\cite{syed20_interspeech}} & - & - & - & \small85.42 & \small-\\ \hline
\small \textit{\cite{pompili20_interspeech}} & \small 94.12 & \small66.67 & \small78.05 & \small81.25 &\small 95.83\\ 
\midrule
\multicolumn{6}{>{\columncolor[gray]{.8}}l}{\textbf{Proposed Transformer-based models}} \\
\textit{\small{BERT}} & 87.19 & 81.66 & 86.73 & 87.50 & 93.33 \\
& $\pm$3.25 & $\pm$5.00 & $\pm$4.53 & $\pm$4.37 & $\pm$5.65\\ \hline
\textit{\small{BioBERT}} & 86.87 & 78.33 & 82.11 & 82.92 & 87.50 \\
& $\pm$6.09 & $\pm$4.86 & $\pm$2.83 & $\pm$3.06 & $\pm$6.97\\ \hline
\textit{\small{BioClinicalBERT}} & 95.03 & 76.66 & 84.72 & 86.25 & 95.83 \\
& $\pm$3.03 & $\pm$4.99 & $\pm$2.74 & $\pm$2.12 & $\pm$2.64\\ \hline
\textit{\small{ConvBERT}} & 83.51 & 79.99 & 81.65 & 82.08 & 84.16 \\
& $\pm$1.23 & $\pm$4.08 & $\pm$2.06 & $\pm$1.66 & $\pm$1.66\\ \hline
\textit{\small{RoBERTa}} & 90.24 & 76.66 & 82.81 & 84.16 & 91.66\\
& $\pm$2.81 & $\pm$4.99 & $\pm$3.52 & $\pm$2.83 & $\pm$2.64\\ \hline
\textit{\small{ALBERT}}& 79.15 & 78.33 & 78.45 & 78.33 & 78.33  \\
& $\pm$7.89 & $\pm$3.11 & $\pm$3.12 & $\pm$3.86 & $\pm$8.89 \\ \hline
\textit{\small{XLNet}} & 85.58 & 68.33 & 75.75 & 78.33 & 88.33  \\
& $\pm$2.77 & $\pm$6.77 & $\pm$4.05 & $\pm$2.82 & $\pm$3.12\\ \midrule
\multicolumn{6}{>{\columncolor[gray]{.8}}l}{\textbf{Proposed Transformer-based models with co-attention mechanism}} \\
\textit{\small{BERT}} & 83.67 & 84.16 & 83.85 & 83.75 & 83.33 \\
\textit{\small{Co-Attention}} & $\pm$3.36 & $\pm$1.66 & $\pm$1.09 & $\pm$1.56 & $\pm$4.56\\ 
\hline
\textit{\small{BioBERT}} & 85.41 & 76.66 & 80.72 & 81.66 & 86.66 \\
\textit{\small{Co-Attention}} & $\pm$4.91 & $\pm$3.33 & $\pm$3.16 & $\pm$3.06 & $\pm$4.86\\ \hline
\textit{\small{BioClinicalBERT}} & 82.60 & 81.66 & 81.99 & 82.08 & 82.50 \\
\textit{\small{Co-Attention}} & $\pm$3.60 & $\pm$4.25 & $\pm$2.11 & $\pm$2.12 & $\pm$4.86\\ \hline
\textit{\small{ConvBERT}} & 83.78 & 81.66 & 82.43 & 82.50 & 83.33 \\
\textit{\small{Co-Attention}} & $\pm$6.13 & $\pm$4.24 & $\pm$2.37 & $\pm$3.12 & $\pm$8.74\\ \hline
\textit{\small{RoBERTa}} & 79.39 & 79.16 & 79.06 & 79.16 & 79.16\\
\textit{\small{Co-Attention}} & $\pm$2.26 & $\pm$6.45 & $\pm$2.15 & $\pm$1.32 & $\pm$4.56\\ \hline
\textit{\small{ALBERT}}& 77.94 & 84.16 & 80.77 & 80.00 & 75.83 \\
\textit{\small{Co-Attention}} & $\pm$3.20 & $\pm$4.86 & $\pm$1.68 & $\pm$1.66 & $\pm$5.53 \\ \hline
\textit{\small{XLNet}} & 85.63 & 69.16 & 76.42 & 78.75 & 88.33 \\
\textit{\small{Co-Attention}} & $\pm$3.45 & $\pm$5.00 & $\pm$3.75 & $\pm$3.06 & $\pm$3.12 \\
\bottomrule
\end{tabular}
\label{compare}
\end{table}

\subsection{Multi-Task Learning Experiments}
\subsubsection{Primary Task}
The results of the introduced models described in Section \ref{mtl_models} are reported in Table \ref{compare_mtl}. Also, Table \ref{compare_mtl} provides a comparison of our introduced approaches with state-of-the-art approaches.

With regards to our introduced models, one can easily observe that MTL-BERT (Multiclass) outperforms MTL-BERT-DE (Multiclass) in terms of all the evaluation metrics except Recall. Specifically, MTL-BERT (Multiclass) surpasses MTL-BERT-DE (Multiclass) in Precision by 3.40\%, in F1-score by 0.88\%, in Accuracy by 1.25\%, and in Specificity by 4.16\%. Although MTL-BERT-DE (Multiclass) surpasses MTL-BERT (Multiclass) in Recall by 1.67\%, MTL-BERT (Multiclass) obtains a higher F1-score, which constitutes the weighted average of Precision and Recall. Therefore, MTL-BERT (Multiclass) constitutes our best performing model in the MTL framework.

In comparison to the research work \cite{10.3389/fcomp.2021.624683}, as one can easily observe, both our introduced models attain a higher Accuracy score. To be more precise, MTL-BERT (Multiclass) outperforms BERT base \cite{10.3389/fcomp.2021.624683} in Accuracy by 5.42\%. In addition,  MTL-BERT-DE (Multiclass) surpasses the research work \cite{10.3389/fcomp.2021.624683} in Accuracy by 4.17\%. These differences in performance are attributable to the fact that we adopt a different training procedure than the one adopted by \cite{10.3389/fcomp.2021.624683}, we consider the MMSE task as a multiclass classification task instead of a regression task, as well as to the different architectures proposed.

\begin{table}[hbt]
\caption{Performance comparison among proposed MTL models and state-of-the-art approaches on the ADReSS Challenge test set for the primary task (AD Classification Task). Reported values are mean $\pm$ standard deviation. Results are averaged across five runs.}
\centering
\begin{tabular}{lccccc}
\toprule
\multicolumn{1}{l}{}&\multicolumn{5}{c}{\textbf{Evaluation metrics}}\\
\cline{2-6} 
\multicolumn{1}{l}{\textbf{Architecture}}&\textbf{Prec.}&\textbf{Rec.}&\textbf{F1-score}&\textbf{Acc.}&\textbf{Spec.}\\
\midrule
\multicolumn{6}{>{\columncolor[gray]{.8}}l}{\textbf{Comparison with state-of-the-art approaches}} \\
\textit{\small{\cite{10.3389/fcomp.2021.624683}}} & - & - & - & 80.83 & -  \\
& - & - & - & $\pm$1.56 & - \\
\midrule
\multicolumn{6}{>{\columncolor[gray]{.8}}l}{\textbf{Proposed Multi-task learning models}} \\
\textit{\small{MTL-BERT}} & 88.59 & 83.33 & 85.84 & 86.25 & 89.16 \\
\textit{\small{(Multiclass)}} & $\pm$3.05 & $\pm$2.64 & $\pm$2.12 & $\pm$2.13 & $\pm$3.33\\ \hline
\textit{\small{MTL-BERT-DE}} & 85.19 & 85.00 & 84.96 & 85.00 & 85.00\\
\textit{\small{(Multiclass)}} & $\pm$3.46 & $\pm$5.00 & $\pm$2.60 & $\pm$2.43 & $\pm$4.25\\
\bottomrule
\end{tabular}
\label{compare_mtl}
\end{table}

\subsubsection{Auxiliary Task}
The results of the introduced models mentioned in Section \ref{mtl_models} for the auxiliary task (MMSE Classification task) are reported in Table \ref{results_auxiliary}. 

As one can easily observe, MTL-BERT (Multiclass) obtains an average weighted Precision of 73.62\% surpassing MTL-BERT-DE (Multiclass) by 3.12\%. However, MTL-BERT-DE (Multiclass) outperforms MTL-BERT (Multiclass) in average weighted Recall and average weighted F1-score by 1.26\% and 3.82\% respectively.

\begin{table}[hbt]
\caption{Results of the proposed MTL models on the ADReSS Challenge test set for the auxiliary task (MMSE Classification Task). Reported values are mean $\pm$ standard deviation. Results are averaged across five runs.}
\centering
\begin{tabular}{lccccc}
\toprule
\multicolumn{1}{l}{}&\multicolumn{3}{c}{\textbf{Evaluation metrics}}\\
\cline{2-4} 
\multicolumn{1}{l}{\textbf{Architecture}}&\textbf{Avg. W. Prec.}&\textbf{Avg. W. Rec.}&\textbf{Avg. W. F1-score} \\
\midrule
\multicolumn{4}{>{\columncolor[gray]{.8}}l}{\textbf{Proposed Multi-task learning models}} \\
\textit{\small{MTL-BERT}} & 73.62 & 69.16 & 64.75 \\
\textit{\small{(Multiclass)}} & $\pm$2.95 & $\pm$4.04 & $\pm$3.50  \\ \hline
\textit{\small{MTL-BERT-DE}} & 70.50 & 70.42 & 68.57 \\
\textit{\small{(Multiclass)}} & $\pm$5.59 & $\pm$3.06 & $\pm$2.04 \\
\bottomrule
\end{tabular}
\label{results_auxiliary}
\end{table}

\section{Analysis of the Language used in Control and Dementia groups}

We finally perform an extensive analysis to uncover some unique characteristics, which discriminate the AD patients from the non-AD ones, and understand the predictions made by our best performing model as well as its limits.

\subsection{Text Statistics}

We first extract some statistics, namely the syllable count, the lexicon count, the difficult words, and the sentence count, using the \texttt{TEXTSTAT} library in Python, in order to understand better the differences in language used between control and dementia groups. More specifically, the syllable count refers to the number of syllables, the lexicon count to the number of words, and the sentence count to the number of sentences present in the given text. With regards to the difficult words, they refer to the number of polysyllabic words with a Syllable Count > 2 that are not included in the list of words of common usage in English \cite{portelli-etal-2021-bert}. After extracting these statistics per transcript, we calculate the mean and standard deviation for both control and dementia groups. We test for statistical significance using an independent t-test for each metric between control and dementia groups and adjust the p-values using Benjamini-Hochberg correction \cite{benjamini1995controlling}. As one can easily observe in Table \ref{textstat}, the control group presents a significantly higher number of syllables, lexicon, and difficult words than the dementia group.

\begin{table}[hbt]
\caption{mean $\pm$ standard deviation metrics per transcript. $\dagger$ indicates statistical significance between transcripts of control and dementia groups. All differences are significant at $p<0.05$ after Benjamini-Hochberg correction.}
\centering
\begin{tabular}{ccc}
\toprule
\multicolumn{1}{c}{} & \multicolumn{2}{c}{Transcript} \\ \cline{2-3}
\textbf{Metric} & Control & Dementia \\ \midrule
Syllable Count$\dagger$ & 151.63 $\pm$ 79.98 & 119.95 $\pm$ 71.18 \\
Lexicon Count$\dagger$ & 107.49 $\pm$ 62.02 & 86.08 $\pm$ 54.10 \\
Difficult Words$\dagger$ & 10.58 $\pm$ 3.64 & 6.38 $\pm$ 3.53 \\
Sentence Count & 1.67 $\pm$ 1.03 & 1.92 $\pm$ 1.62 \\ \bottomrule
\end{tabular}
\label{textstat}
\end{table}

\subsection{Vocabulary Uniqueness}

In order to understand the vocabulary similarities and differences between control and dementia groups, we
adopt the methodology proposed by \cite{rissola2020beyond}. Formally, let $\mathcal{P}$ and $\mathcal{C}$ be the sets of unique words included in the control group and dementia group respectively. Next, we calculate the Jaccard's index given by \eqref{jaccard}, in order to measure the similarity between finite sample sets. More specifically, the Jaccard's index is a number between 0 and 1, where 1 indicates that the two sets, namely $\mathcal{P}$ and $\mathcal{C}$, have the same elements, while 0 indicates that the two sets are completely different. 

\begin{equation}
    J(P,C) = |P \cap C|/|P \cup C|
\label{jaccard}
\end{equation}

As observed in Table \ref{jaccard_index}, the Jaccard's index between the control and dementia groups is equal to 0.4049, which indicates that people with dementia tend to use a different vocabulary than those in the control group.

\begin{table}[hbt]
\caption{Jaccard's Index between transcripts of control and dementia group}
\centering
\begin{tabular}{cc}
\toprule
\textbf{Jaccard's Index between transcripts} & \textbf{Result} \\ \midrule
\textit{J}($\mathcal{P}$= control, $\mathcal{C}$=dementia) & 0.4049 \\  \bottomrule
\end{tabular}
\label{jaccard_index}
\end{table}

\subsection{Word Usage}

Apart from finding the vocabulary similarities and differences, it is imperative that patterns of word usage be investigated. Thus, following the methodology introduced in \cite{rissola2020beyond}, the main objective of this section is to explore the differences between the two classes (control and dementia) with regard to the probability of using specific words more than others. Formally, let $D_1$ and $D_2$ be two documents, where $D_1$ includes all the transcriptions of the control group, whereas $D_2$ consists of transcriptions of the dementia group. Moreover, we define $S$ as the entire corpus consisting of $D_1$ and $D_2$. Now we can define the probability of a word $w_i$ in the document $D_1$ in a collection of documents $S$ given by \eqref{p_word_1}:

\begin{equation}
    P(w_i |D_1, S)= (1-\alpha_D)P(w_i |D_1) + \alpha_D P(w_i |S)
\label{p_word_1}
\end{equation}

Similarly, we can define the probability of a word $w_i$ in the document $D_2$ in a collection of documents $S$ given by \eqref{p_word_2}:

\begin{equation}
    P(w_i |D_2, S)= (1-\alpha_D)P(w_i |D_2) + \alpha_D P(w_i |S)
\label{p_word_2}
\end{equation}

We employ the Jelinek-Mercer smoothing method and consider that $\alpha_D \in [0,1]$. More specifically, $\alpha_D$ is a parameter that controls the probability of words included only in one document ($D_1$ or $D_2$). In our experiments, we set $\alpha_D$ equal to 0.2. 

Moreover, we define $P(w_i |S) = \frac{s_{w_i}}{|S|}$, where $s_{w_i}$ denotes the number of times a word $w_i$ is included in the collection, whereas $|S|$ is the total number of words occurrences in the collection. Similarly, $P(w_i |D_1) = \frac{d_{w_i}}{|D_1|}$, where $d_{w_i}$ denotes the number of times a word $w_i$ is presented in the document $D_1$, whereas $|D_1|$ is the total number of words occurrences in the document $D_1$. The same methodology has been adopted for calculating the $P(w_i |D_2)$.

After having calculated the two distributions, i.e., $P(w_i |D_1, S)$ and $P(w_i |D_2, S)$, we exploit the Kullback-Leibler (KL) divergence, in order to measure the difference of these two distributions. KL-divergence is always greater than zero and is given by \eqref{kl-divergence}. The larger it gets, the more different the two distributions are.

\begin{equation}
    KL(P||C)= \sum_x P(x)log \frac{P(x)}{C(x)}
\label{kl-divergence}
\end{equation}

As one can easily observe in Table \ref{kldivergence}, the KL divergence between control and dementia groups is high indicating that these two groups present differences regarding the probability of using some words more than others. Our findings agree with the ones in \cite{rissola2020beyond}, where the authors state that there are clear differences in terms of language use between positive (depression and self-harm) and control group, where the values of KL-divergence range from 0.18 to 0.21.
\begin{table}[hbt]
\caption{Kullback-Leibler divergence}
\centering
\begin{tabular}{cc}
\toprule
\textbf{KL divergence} & \textbf{Result} \\ \midrule
KL(Control \(\vert \vert\) Dementia) & 0.2047 \\ \hline
KL(Dementia \(\vert \vert\) Control) & 0.2161 \\ \bottomrule
\end{tabular}
\label{kldivergence}
\end{table}

\subsection{Linguistic Feature Analysis} \label{explain_features}

Following the method introduced by \cite{schwartz2013personality}, the main objective of this section is to shed light on which unigrams and pos-tags are mostly correlated with each class separately. To facilitate this, we compute the point-biserial correlation between each feature (unigram and pos-tag) across all the transcriptions and a binary label (0 for the control and 1 for the dementia group). Before computing the correlation, we normalize features so that they sum up to 1 across each transcription. We use the point-biserial correlation, since it is a correlation used between continuous and binary variables. It returns a value between -1 and 1. Since we are only interested in the strength of the correlation, we compute the absolute value, where negative correlations refer to the control group (label 0) and positive correlations refer to the dementia one (label 1). We report our findings in Table \ref{correlations}, where all correlations are significant at $p<0.05$, with Benjamini-Hochberg correction \cite{benjamini1995controlling} for multiple comparisons.

As one can easily observe, the pos-tags associated with the dementia group are the following: RB (adverbs), PRP (personal pronoun), VBD (verb in past tense), and UH (interjection). On the other hand, people in the control group tend to use VBG (verb, gerund, or present participle), DT (determiner), and NN (noun). These findings can be justified in Table \ref{feature_analysis}, where we present three examples of transcripts belonging to the control group and three examples of transcripts belonging to the dementia one. More specifically, we have assigned colours to different pos-tags, so as to render the differences in the language patterns used by each group easily understandable to the reader. To be more precise, \textcolor{red}{red} colour indicates the VBG pos-tag, \textcolor{yellow}{yellow} refers to the DT pos-tag, \textcolor{Fuchsia}{fuchsia} to the RB pos-tag, \textcolor{Apricot}{apricot} to the PRP pos-tag, \textcolor{NavyBlue}{navy blue} to the VBD pos-tag, and the \textcolor{PineGreen}{pine green} to the UH pos-tag. 

We observe that people in the dementia group tend to use personal pronouns (he, she, I, them etc.) very often, since they are unable to remember the specific terms (mom, boy, etc.). This finding agrees with the research conducted by \cite{almor1999alzheimer}, where the authors state that personal pronouns present a high frequency in the speech of AD patients, since these people cannot find the target word. To be more precise, in a conversation people have to remember what they have said during the entire conversation. However, this is not possible in AD patients, who present working memory impairment and thus tend to produce empty conversational speech (use of personal pronouns). On the other hand, people in the control group tend to use more nouns instead of personal pronouns, since they are able to maintain various kinds of information. 

Moreover, AD patients tend to use verbs in the past tense (were, forgot, did, started) in contrast to people who are not suffering from dementia and use verbs in the present participle. One typical example that can illustrate this difference can be seen in the fifth transcription in Table \ref{feature_analysis}, i.e., \textit{"oh have you heard of that new game that they started to play after christmas ? did you"}. The AD patient perhaps remembers a personal story from the past that wants to narrate, instead of the task he has been assigned to conduct. Therefore, the patient is not able to stay focused on describing the picture. This finding is consistent with \cite{watson1999analysis,garcia1997analysis}, where the authors state that AD patients present difficulty in maintaining and continuing the development of a topic and thus demonstrate unexpected topic shifts. Also, this finding reveals a difference in language used by the AD patients and the agrammatic aphasics. Specifically, patients with agrammatic aphasia typically have problems using past tense inflection and instead rely on infinitive or present tense verb forms \cite{doi:10.3109/02699206.2012.751626}.

In addition, AD patients tend to use the UH (oh, yeah, well) and the RB (maybe, probably) pos-tags, since they are not certain of what they are describing due to the cognitive impairment. Concurrently, the UH pos-tag constitutes an example of empty speech. More specifically, this pos-tag is used as filler at the beginning of each utterance, since AD patients are thinking of what to say.

\begin{table}[hbt]
\caption{Features associated with control and dementia subjects, sorted by point-biserial correlation. All correlations are significant at \textit{p} < 0.05 after Benjamini-Hochberg correction.}
\centering
\begin{tabular}{|c|c||c|c|}
\hline
\multicolumn{2}{|c||}{\textbf{Control}}&\multicolumn{2}{c|}{\textbf{Dementia}}\\
\hline \hline
\textbf{Unigrams} & \textbf{corr.} & \textbf{Unigrams} & \textbf{corr.}\\ \hline
is & 0.364 & here & 0.310 \\ \hline
curtains & 0.361 & - & - \\ \hline
window & 0.301 & - & -\\ \hline
are & 0.300 & - & -\\ \hline
\hline \hline
\textbf{POS} & \textbf{corr.} & \textbf{POS} & \textbf{corr.}\\ \hline
VBG & 0.285 & RB & 0.388 \\ \hline
DT & 0.216 & PRP & 0.354 \\ \hline
NN & 0.210 & VBD & 0.275 \\ \hline
- & - & UH & 0.242 \\ \hline
\end{tabular}
\label{correlations}
\end{table}

\begin{table*}
\caption{Examples of transcripts along with their labels. \textcolor{red}{red} colour indicates the VBG pos-tag, \textcolor{yellow}{yellow} refers to the DT pos-tag, \textcolor{Fuchsia}{fuchsia} to the RB pos-tag, \textcolor{Apricot}{apricot} to the PRP pos-tag, \textcolor{NavyBlue}{navy blue} to the VBD pos-tag, and the \textcolor{PineGreen}{pine green} to the UH pos-tag.}
\begin{tabularx}{\textwidth}{X>{\hsize=.07\hsize}X}
\toprule
\multicolumn{1}{c}{\textbf{Transcript}} & \multicolumn{1}{c}{\textbf{Label}} \\ \midrule
\textit{" well \textcolor{yellow}{the} girl is \textcolor{red}{watching} \textcolor{yellow}{the} boy go into \textcolor{yellow}{the} cookie jar . he has \textcolor{yellow}{a} cookie in his hand . he's on \textcolor{yellow}{the} stool . \textcolor{yellow}{the} stool is \textcolor{red}{falling} . \textcolor{yellow}{the} mother is \textcolor{red}{drying} dishes . has \textcolor{yellow}{a} plate in her hand . sink is \textcolor{red}{overflowing} . there's water on \textcolor{yellow}{the} floor . she's \textcolor{red}{stepping} in \textcolor{yellow}{the} water . something that's \textcolor{red}{going} on you said ? \textcolor{yellow}{the} little girl looks like she's \textcolor{red}{motioning} to \textcolor{yellow}{the} boy to be quiet . and I don't know what else . \textcolor{yellow}{the} woman's \textcolor{red}{looking} out \textcolor{yellow}{the} window . \textcolor{yellow}{the} window's open . "} & Control \\ \midrule
\textit{" action . alright . \textcolor{yellow}{a} lady's \textcolor{red}{drying} dishes . \textcolor{yellow}{the} boy was standing on \textcolor{yellow}{a} stool but \textcolor{yellow}{the} action is that \textcolor{yellow}{the} stool has slipped and he is \textcolor{red}{falling} . and \textcolor{yellow}{the} girl has her hand raised \textcolor{red}{reaching} for \textcolor{yellow}{a} cookie . and there's a lot of action in \textcolor{yellow}{the} sink here . \textcolor{yellow}{the} water is \textcolor{red}{flowing} out . she is apparently so \textcolor{red}{daydreaming} that she doesn't realize that \textcolor{yellow}{the} sink is \textcolor{red}{overflowing} . any more action ? or is that enough action ? "} & Control \\ \midrule
\textit{" \textcolor{red}{touching} lip . \textcolor{red}{raising} arm . is that what you mean ? \textcolor{red}{reaching} for cookie . \textcolor{red}{handing} cookie down . \textcolor{red}{slipping} from stool . stool \textcolor{red}{falling} over . \textcolor{red}{wiping} dishes . water \textcolor{red}{running} . water \textcolor{red}{overflowing} . breeze . I don't know if that's action . \textcolor{red}{stepping} out from water . I guess that's it . "} & Control \\ \midrule
\textit{" alright . \textcolor{Apricot}{I} see the little boy stealing cookies from the cookie jar . and \textcolor{Apricot}{he} \textcolor{NavyBlue}{gave} some to the little girl and \textcolor{Apricot}{she}'s eating some of the cookies . and \textcolor{Apricot}{I} guess this is mama and \textcolor{Apricot}{she}'s washing the dishes . and \textcolor{Apricot}{she} \textcolor{NavyBlue}{dropped} a dish . no \textcolor{Apricot}{she} \textcolor{NavyBlue}{didn't} drop a dish . the water that \textcolor{Apricot}{she}'s washing the dishes with \textcolor{Apricot}{she} let run . and \textcolor{Apricot}{it}'s overflown . that doesn't sound \textcolor{Fuchsia}{right} . \textcolor{NavyBlue}{did} \textcolor{Apricot}{it} ? \textcolor{Apricot}{we} \textcolor{NavyBlue}{forgot} to turn off the spigot . and so the water is running off onto the floor \textcolor{Fuchsia}{here} . and mom \textcolor{Fuchsia}{apparently} is washing the dishes . and \textcolor{Fuchsia}{here}'s this little boy stealing the cookies . \textcolor{Apricot}{he}'s gonna fall because the four legged stool is gonna fall over with \textcolor{Apricot}{him} and the cookie jar . and mama's drying the dishes as usual for mamas if \textcolor{Apricot}{they} don't have a husband that dries \textcolor{Apricot}{them} or washes \textcolor{Apricot}{them} or whatever . let's see \textcolor{Fuchsia}{now} . \textcolor{Apricot}{I} guess there's more things \textcolor{Apricot}{I}'m sposta see . let's see \textcolor{Fuchsia}{here now} . \textcolor{PineGreen}{oh} and the water is flowing out of the sink they \textcolor{NavyBlue}{forgot} to turn off whoever's doing the dishwashing . mom \textcolor{Fuchsia}{apparently here} , \textcolor{Apricot}{she} \textcolor{NavyBlue}{forgot} to turn off the water and the water is spilling out onto the kitchen floor . and the little girl has pushed over the stool with the boy that \textcolor{NavyBlue}{was} reaching up to get the cookies . either \textcolor{Apricot}{she} \textcolor{NavyBlue}{pushed} it over or \textcolor{Apricot}{he} \textcolor{NavyBlue}{fell} over with it . \textcolor{Apricot}{you} know it excuse me but \textcolor{Apricot}{you} know \textcolor{Apricot}{I} \textcolor{NavyBlue}{was} ... "} & Dementia \\ \midrule
\textit{" mhm . \textcolor{PineGreen}{oh} \textcolor{Apricot}{I} see a part of the whole kitchen . is that all the kitchen or isn't \textcolor{Apricot}{it} ? \textcolor{PineGreen}{oh} \textcolor{Apricot}{I} can't read ... a lady a mother \textcolor{NavyBlue}{were} in her kitchen . in her kitchen doing some work \textcolor{Apricot}{I} suppose . and there's another woman there sharing their pleasures or whatever . \textcolor{PineGreen}{oh} have \textcolor{Apricot}{you} heard of that new game that \textcolor{Apricot}{they} \textcolor{NavyBlue}{started} to play after christmas ? \textcolor{NavyBlue}{did} \textcolor{Apricot}{you} ? is a . \textcolor{PineGreen}{well} \textcolor{Apricot}{it} looks like ... \textcolor{Apricot}{I}'d say this is ... \textcolor{PineGreen}{well} let's see . \textcolor{Apricot}{it} looks like ... \textcolor{PineGreen}{oh} ... . my wife will beat \textcolor{Apricot}{me} by a couple rows of this . that's like the washing machine ? or let \textcolor{Apricot}{me} see . \textcolor{Apricot}{I} can't ... \textcolor{PineGreen}{oh} that's the son come from school \textcolor{Fuchsia}{maybe} or something . that's a youngster there . \textcolor{PineGreen}{well} that's \textcolor{Fuchsia}{just} as though \textcolor{Apricot}{they} getting ready to go to school or \textcolor{Apricot}{they}'re \textcolor{Fuchsia}{just} coming out from school . and right there \textcolor{Apricot}{he}'s same as back there except for \textcolor{Fuchsia}{down there} in the bottom \textcolor{Apricot}{I} think \textcolor{Apricot}{it}'s ... that's a little . "} & Dementia \\ \midrule
\textit{" yes . the water ? \textcolor{PineGreen}{well} let's see . there's something hasta be where the water goes \textcolor{Fuchsia}{down over} . there's \textcolor{Fuchsia}{probably} something that's ... or \textcolor{Apricot}{they} don't have it open or something might have. \textcolor{Apricot}{I} don't know . what ..? when the water goes \textcolor{Fuchsia}{down} what do you call that ? this \textcolor{Fuchsia}{here} . right \textcolor{Fuchsia}{here} . this . what do you call that ? what is that ? what is that ? \textcolor{Apricot}{I} don't know ! that's what \textcolor{Apricot}{I}'m saying . \textcolor{Apricot}{I} don't know what that is . the what ? a pipe . \textcolor{PineGreen}{oh} water pipe ! \textcolor{PineGreen}{oh yeah} . okay . \textcolor{PineGreen}{well} then \textcolor{Fuchsia}{maybe} the water pipe is not broke but there must be things in there . that the water will not go \textcolor{Fuchsia}{down} . \textcolor{Apricot}{I} don't know . huh ? what's happening to the water ? \textcolor{PineGreen}{well} the water is going \textcolor{Fuchsia}{down} in the ... \textcolor{Apricot}{I} don't know . what would you call this ? floor ! \textcolor{PineGreen}{yeah} okay . \textcolor{PineGreen}{yeah} . \textcolor{PineGreen}{well} down on this side of the picture . \textcolor{PineGreen}{well} this thing \textcolor{Fuchsia}{here} is turning over . \textcolor{PineGreen}{yeah} . no , uhuh . \textcolor{Apricot}{I} don't know what's going on . \textcolor{PineGreen}{well} \textcolor{Apricot}{he}'s \textcolor{Fuchsia}{probably} getting ... what's this \textcolor{Fuchsia}{here} ? cocoa jar ? what's this cocoa ? c o o k i e . \textcolor{Apricot}{I} don't know . \textcolor{Apricot}{I} don't know what ..? huh ? cookie , \textcolor{PineGreen}{oh} a cookie . \textcolor{PineGreen}{oh} ! \textcolor{PineGreen}{oh} okay . mhm . \textcolor{PineGreen}{well} \textcolor{Apricot}{he}'s getting it out . and \textcolor{Apricot}{he}'s gonna give it to the girl /. \textcolor{Fuchsia}{down} \textcolor{Fuchsia}{here} . mhm . going on in the picture ? \textcolor{PineGreen}{well} the boy is giving her the girl the cookie . this \textcolor{Fuchsia}{probably} is broke . so the water will not go \textcolor{Fuchsia}{down} in and \textcolor{Apricot}{it}'s coming up and going in \textcolor{Fuchsia}{here} huh . \textcolor{PineGreen}{well} \textcolor{Apricot}{it} looks like \textcolor{Apricot}{she} \textcolor{NavyBlue}{was} gonna wash . what \textcolor{Apricot}{they} eat with , all that . what do \textcolor{Apricot}{you} call that ? what do \textcolor{Apricot}{you} call this ? a plate ? \textcolor{PineGreen}{oh yeah} . what \textcolor{Apricot}{you} eat on . is that what \textcolor{Apricot}{you} call them a plate ? \textcolor{PineGreen}{oh} this is a cup ? \textcolor{PineGreen}{oh} \textcolor{Fuchsia}{maybe} , \textcolor{Apricot}{I} don't know . mhm . okay . "} & Dementia \\ 
\bottomrule
\end{tabularx}
\label{feature_analysis}
\end{table*}

\subsection{Explainability - Error Analysis}

In this section, we employ LIME \cite{ribeiro2016should} (using 5000 samples) to explain the predictions made by our best performing model, namely BERT, and shed more light regarding the differences in language between AD and non-AD patients. More specifically, LIME generates local explanations for any machine learning classifier by introducing an interpretable model, which is trained on data generated through observing differences in the classification performance when removing tokens from the input string.

\begin{figure*}[hbt]
\centering
\subfloat[]{\label{label_d_pred_c_1}\includegraphics[width=1\textwidth,height=13mm]{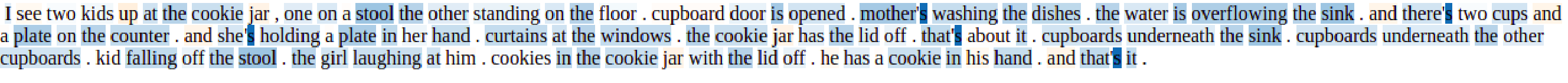} } \\

\subfloat[]{\label{label_d_pred_c_2}\includegraphics[width=1\textwidth,height=13mm]{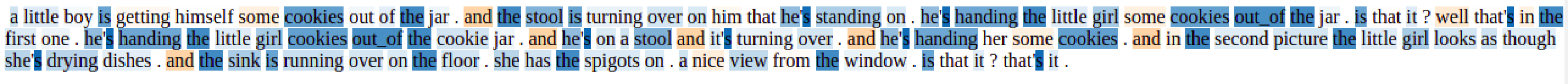}}
\caption{Label: Dementia, Prediction: Control}\label{fig:a}
\end{figure*}

\begin{figure*}[hbt]
\centering
\subfloat[]{\includegraphics[width=1\textwidth,height=13mm]{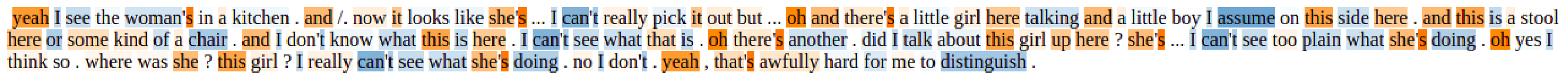}
\label{label_d_pred_d_1}}

\centering
\subfloat[]{\includegraphics[width=1\textwidth,height=15mm]{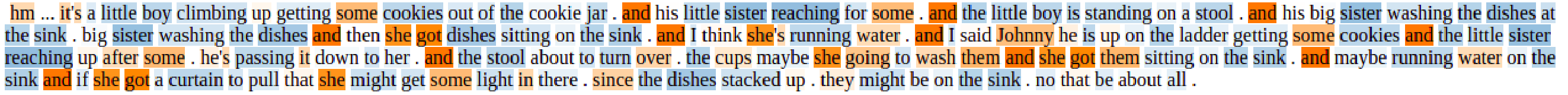}
\label{label_d_pred_d_2}}

\centering
\subfloat[]{\includegraphics[width=1\textwidth,height=12mm]{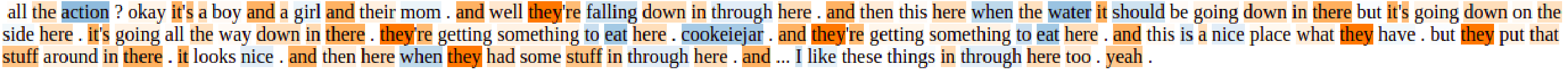}
\label{label_d_pred_d_3}}
\caption{Label: Dementia, Prediction: Dementia}\label{fig:b}
\end{figure*}

\begin{figure*}[hbt]
\centering
\subfloat[]{\includegraphics[width=1\textwidth,height=21mm]{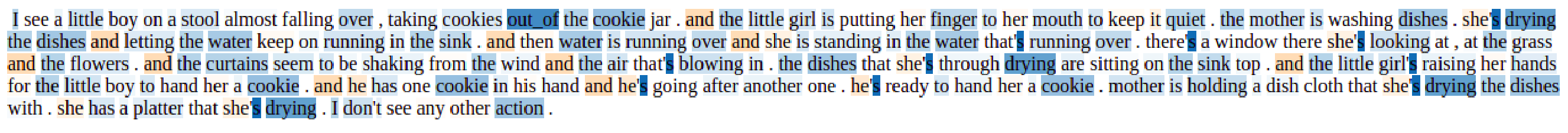}
\label{label_c_pred_c_1}}

\centering
\subfloat[]{\includegraphics[width=1\textwidth,height=15mm]{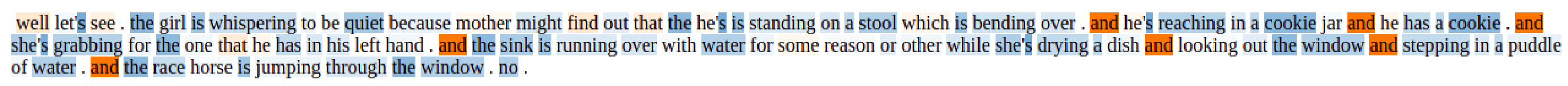}
\label{label_c_pred_c_2}}
\caption{Label: Control, Prediction: Control}\label{fig:c}
\end{figure*}

\begin{figure*}[hbt]
\centering
\subfloat[]{\includegraphics[width=1\textwidth,height=19mm]{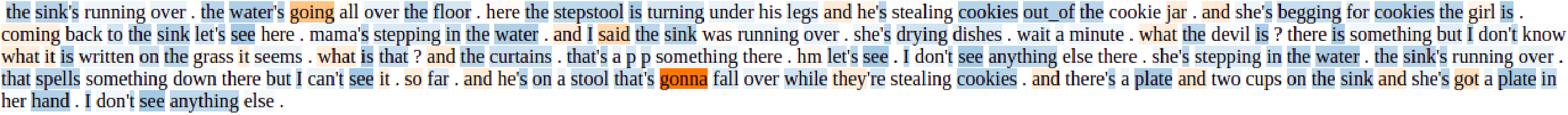}
\label{label_c_pred_d_1}}

\centering
\subfloat[]{\includegraphics[width=1\textwidth,height=17mm]{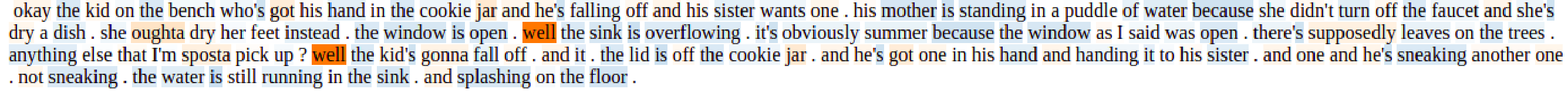}
\label{label_c_pred_d_2}}

\centering
\subfloat[]{\includegraphics[width=1\textwidth,height=10mm]{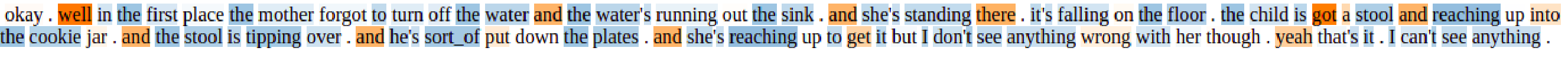}
\label{label_c_pred_d_3}}
\caption{Label: Control, Prediction: Dementia}\label{fig:d}
\end{figure*}

Examples of explanations generated by LIME are illustrated in Figs. \ref{fig:a}-\ref{fig:d}. More specifically, Fig. \ref{fig:a} illustrates two transcripts, whose ground-truth label is dementia, while our model predicts them as belonging to non-AD patients. Fig. \ref{fig:b} refers to transcripts with both ground-truth label and prediction corresponding to dementia. In Fig. \ref{fig:c}, two transcripts are presented, whose prediction is control and true label is control too. Finally, Fig. \ref{fig:d} illustrates transcripts, which are misclassified. The ground-truth is control, whereas the prediction is dementia. Moreover, as one can observe, each token has been assigned a colour, either blue or orange. To be more precise, the blue colour indicates which tokens are indicative of the control group, whilst the orange colour indicates tokens, which are used mainly by AD patients. The more intense the colours are, the more important these tokens are towards the final classification of the transcript.

As one can easily observe in Fig. \ref{fig:b}, tokens belonging to the UH pos-tag, such as yeah and oh, are identified as important for the dementia class by our best performing model. Moreover, personal pronouns (she, they) and verbs in the past tense (got, had) are also indicative of dementia. Also, our model considers the token "here", which corresponds to the RB pos-tag, indicative of the dementia class. These findings are consistent with the ones in Section \ref{explain_features}, where we have found that PRP, VBD, UH pos-tags as well as the unigram "here" are significantly correlated with the dementia class. In addition, our model identifies the repetition of token "and" as important for the dementia class. This finding agrees with previous research works \cite{karlekar-etal-2018-detecting}, where the word "and" indicates a short answer and burst of speech. 

Regarding Fig. \ref{fig:c}, one can easily observe that our model identifies tokens belonging to the VBG (putting, drying, blowing, standing, etc.), DT (the, a), and NN (cookie, action, stool, etc.) pos-tags as significant for the control class. Concurrently, in consistence with the findings in Section \ref{explain_features}, the unigrams "curtain" and "window" are used mainly by non-AD patients. 

With regards to Figs. \ref{fig:a} and \ref{fig:d}, our model is not able to classify these transcripts correctly. One possible reason for such misclassifications has to do with the fact that these transcripts include pos-tags which are indicative of both the control and the dementia class. To be more precise, in Fig. \ref{fig:a}, the majority of tokens in both transcripts belong to the VBG, NN, and DT pos-tags, which are correctly identified by our model as significant for the control group. Words, like "and", "him", and "well" are used in a low frequency. Similarly to Fig. \ref{fig:a}, in Fig. \ref{fig:d}, the majority of tokens in each transcript belong to the pos-tags which are significantly correlated with the dementia class. This can be illustrated in Fig. \ref{label_c_pred_d_3}, where we observe the usage of words, like "and", "yeah", "well" \& "got". 

\section{Conclusions and Future Work}
We introduced both single-task and multi-task learning models. Regarding single-task learning models, we employed several transformer-based networks and compared their performances. Results showed that BERT achieved the highest classification performance with accuracy accounting for 87.50\%. Concurrently, we introduced siamese networks coupled with a co-attention mechanism which can detect AD patients with an accuracy up to 83.75\%. In terms of the multi-task learning setting, it consisted of two tasks, the primary and the auxiliary one. The primary task was the identification of dementia (binary classification), whereas the auxiliary task was the categorization of the severity of dementia into one of the four categories -healthy, mild/moderate/severe dementia- (multiclass classification). Specifically, we proposed two multi-task learning models. Results showed that our model achieves competitive results in the MTL framework reaching accuracy up to 86.25\% on the detection of AD patients. Next, we performed an in-depth linguistic analysis, in order to understand better the differences in language between AD and non-AD patients. Finally, we employed LIME, in order to shed light on how our best performing model works. Findings suggest that AD patients tend to use personal pronouns, interjection, adverbs, verbs in the past tense, and the token "and" at the beginning of utterances in a high frequency. On the contrary, healthy people use verbs in present participle or gerund, nouns as well as determiners.

One limitation of the current research work is pertinent to the small dataset used for conducting our experiments. However, we opted for this dataset, in order to mitigate different kinds of biases that could otherwise influence the validity of the proposed approaches.

We conducted our experiments on the ADReSS Challenge dataset, which is matched for gender and age and consists of a statistically balanced, acoustically enhanced set of recordings of spontaneous speech. Therefore, the results of this study could be integrated into an application, which will predict whether a person is an AD patient and will provide at the same time the reasons for this prediction via the explainability method. 

In the future, we plan to investigate multimodal deep learning models incorporating both text and audio. Specifically, we plan to propose end-to-end trainable deep neural networks in contrast to existing research initiatives, which train multiple models separately and then use majority-voting approaches. In addition, our aim is to investigate fusion methods, in order to assign more importance to the most relevant modality and suppress the irrelevant information. Another future plan is to exploit further explainability techniques, such as anchor explanations \cite{Ribeiro_Singh_Guestrin_2018}.

\bibliographystyle{unsrt}  
\bibliography{references}  

\begin{thebibliography}{10}

\bibitem{WinNTT}
Alzheimer’s~Association (2021).
\newblock \textit{What Is Dementia? Alzheimer’s Disease and Dementia.}
\newblock Available online at:
  \url{https://www.https://www.alz.org/alzheimers-dementia/what-is-dementia}.
\newblock Accessed: 2021-07-30.

\bibitem{ZHANG2019185}
Fan Zhang, Zhenzhen Li, Boyan Zhang, Haishun Du, Binjie Wang, and Xinhong
  Zhang.
\newblock Multi-modal deep learning model for auxiliary diagnosis of
  alzheimer’s disease.
\newblock {\em Neurocomputing}, 361:185--195, 2019.

\bibitem{hassan2017machine}
Syed~Asif Hassan and Tabrej Khan.
\newblock A machine learning model to predict the onset of alzheimer disease
  using potential cerebrospinal fluid (csf) biomarkers.
\newblock {\em International Journal of Advanced Computer Science and
  Applications}, 8(12):124--131, 2017.

\bibitem{DAVATZIKOS20112322.e19}
Christos Davatzikos, Priyanka Bhatt, Leslie~M. Shaw, Kayhan~N. Batmanghelich,
  and John~Q. Trojanowski.
\newblock Prediction of mci to ad conversion, via mri, csf biomarkers, and
  pattern classification.
\newblock {\em Neurobiology of Aging}, 32(12):2322.e19--2322.e27, 2011.

\bibitem{IERACITANO2020176}
Cosimo Ieracitano, Nadia Mammone, Amir Hussain, and Francesco~C. Morabito.
\newblock A novel multi-modal machine learning based approach for automatic
  classification of eeg recordings in dementia.
\newblock {\em Neural Networks}, 123:176--190, 2020.

\bibitem{bib:LuzHaiderEtAl20ADReSS}
Saturnino Luz, Fasih Haider, Sofia de~la Fuente, Davida Fromm, and Brian
  MacWhinney.
\newblock {Alzheimer's} dementia recognition through spontaneous speech: The
  {ADReSS Challenge}.
\newblock In {\em Proceedings of INTERSPEECH 2020}, Shanghai, China, 2020.

\bibitem{luz21_interspeech}
Saturnino Luz, Fasih Haider, Sofia de~la Fuente, Davida Fromm, and Brian
  MacWhinney.
\newblock {Detecting Cognitive Decline Using Speech Only: The ADReSSo
  Challenge}.
\newblock In {\em Proc. Interspeech 2021}, pages 3780--3784, 2021.

\bibitem{weiner2018selecting}
Jochen Weiner and Tanja Schultz.
\newblock Selecting features for automatic screening for dementia based on
  speech.
\newblock In {\em International Conference on Speech and Computer}, pages
  747--756. Springer, 2018.

\bibitem{CALZA2021101113}
Laura Calzà, Gloria Gagliardi, Rema {Rossini Favretti}, and Fabio Tamburini.
\newblock Linguistic features and automatic classifiers for identifying mild
  cognitive impairment and dementia.
\newblock {\em Computer Speech \& Language}, 65:101113, 2021.

\bibitem{10.3389/fnagi.2019.00205}
Kathleen~C. Fraser, Kristina Lundholm~Fors, Marie Eckerström, Fredrik Öhman,
  and Dimitrios Kokkinakis.
\newblock Predicting mci status from multimodal language data using cascaded
  classifiers.
\newblock {\em Frontiers in Aging Neuroscience}, 11:205, 2019.

\bibitem{10.3389/fcomp.2021.640669}
Shamila Nasreen, Morteza Rohanian, Julian Hough, and Matthew Purver.
\newblock Alzheimer’s dementia recognition from spontaneous speech using
  disfluency and interactional features.
\newblock {\em Frontiers in Computer Science}, 3:49, 2021.

\bibitem{khodabakhsh2014natural}
Ali Khodabakhsh, Serhan Ku{\c{s}}xuo{\u{g}}lu, and Cenk Demiro{\u{g}}lu.
\newblock Natural language features for detection of alzheimer's disease in
  conversational speech.
\newblock In {\em IEEE-EMBS International Conference on Biomedical and Health
  Informatics (BHI)}, pages 581--584. IEEE, 2014.

\bibitem{chen2019attention}
Jun Chen, Ji~Zhu, and Jieping Ye.
\newblock An attention-based hybrid network for automatic detection of
  alzheimer's disease from narrative speech.
\newblock In {\em INTERSPEECH}, pages 4085--4089, 2019.

\bibitem{di-palo-parde-2019-enriching}
Flavio Di~Palo and Natalie Parde.
\newblock Enriching neural models with targeted features for dementia
  detection.
\newblock In {\em Proceedings of the 57th Annual Meeting of the Association for
  Computational Linguistics: Student Research Workshop}, pages 302--308,
  Florence, Italy, July 2019. Association for Computational Linguistics.

\bibitem{syed2021automated}
Zafi~Sherhan Syed, Muhammad Shehram~Shah Syed, Margaret Lech, and Elena
  Pirogova.
\newblock Automated recognition of alzheimer’s dementia using
  bag-of-deep-features and model ensembling.
\newblock {\em IEEE Access}, 9:88377--88390, 2021.

\bibitem{10.3389/fcomp.2021.624683}
Youxiang Zhu, Xiaohui Liang, John~A. Batsis, and Robert~M. Roth.
\newblock Exploring deep transfer learning techniques for alzheimer's dementia
  detection.
\newblock {\em Frontiers in Computer Science}, 3:22, 2021.

\bibitem{karlekar-etal-2018-detecting}
Sweta Karlekar, Tong Niu, and Mohit Bansal.
\newblock Detecting linguistic characteristics of {A}lzheimer{'}s dementia by
  interpreting neural models.
\newblock In {\em Proceedings of the 2018 Conference of the North {A}merican
  Chapter of the Association for Computational Linguistics: Human Language
  Technologies, Volume 2 (Short Papers)}, pages 701--707, New Orleans,
  Louisiana, June 2018. Association for Computational Linguistics.

\bibitem{ribeiro2016should}
Marco~Tulio Ribeiro, Sameer Singh, and Carlos Guestrin.
\newblock " why should i trust you?" explaining the predictions of any
  classifier.
\newblock In {\em Proceedings of the 22nd ACM SIGKDD international conference
  on knowledge discovery and data mining}, pages 1135--1144, 2016.

\bibitem{10.3389/fpsyg.2020.624137}
R'mani Haulcy and James Glass.
\newblock Classifying alzheimer's disease using audio and text-based
  representations of speech.
\newblock {\em Frontiers in Psychology}, 11:3833, 2021.

\bibitem{10.3389/fcomp.2021.624659}
Zehra Shah, Jeffrey Sawalha, Mashrura Tasnim, Shi-ang Qi, Eleni Stroulia, and
  Russell Greiner.
\newblock Learning language and acoustic models for identifying alzheimer’s
  dementia from speech.
\newblock {\em Frontiers in Computer Science}, 3:4, 2021.

\bibitem{syed20_interspeech}
Muhammad Shehram~Shah Syed, Zafi~Sherhan Syed, Margaret Lech, and Elena
  Pirogova.
\newblock {Automated Screening for Alzheimer’s Dementia Through Spontaneous
  Speech}.
\newblock In {\em Proc. Interspeech 2020}, pages 2222--2226, 2020.

\bibitem{pompili20_interspeech}
Anna Pompili, Thomas Rolland, and Alberto Abad.
\newblock {The INESC-ID Multi-Modal System for the ADReSS 2020 Challenge}.
\newblock In {\em Proc. Interspeech 2020}, pages 2202--2206, 2020.

\bibitem{pan2019automatic}
Yilin Pan, Bahman Mirheidari, Markus Reuber, Annalena Venneri, Daniel
  Blackburn, and Heidi Christensen.
\newblock Automatic hierarchical attention neural network for detecting ad.
\newblock In {\em Interspeech}, pages 4105--4109, 2019.

\bibitem{kong2019neural}
Weirui Kong, Hyeju Jang, Giuseppe Carenini, and Thalia Field.
\newblock A neural model for predicting dementia from language.
\newblock In {\em Machine Learning for Healthcare Conference}, pages 270--286.
  PMLR, 2019.

\bibitem{pan2021multi}
Yilin Pan, Venkata~Srikanth Nallanthighal, Daniel Blackburn, Heidi Christensen,
  and Aki H{\"a}rm{\"a}.
\newblock Multi-task estimation of age and cognitive decline from speech.
\newblock In {\em ICASSP 2021-2021 IEEE International Conference on Acoustics,
  Speech and Signal Processing (ICASSP)}, pages 7258--7262. IEEE, 2021.

\bibitem{balagopalan20_interspeech}
Aparna Balagopalan, Benjamin Eyre, Frank Rudzicz, and Jekaterina Novikova.
\newblock {To BERT or not to BERT: Comparing Speech and Language-Based
  Approaches for Alzheimer’s Disease Detection}.
\newblock In {\em Proc. Interspeech 2020}, pages 2167--2171, 2020.

\bibitem{10.3389/fcomp.2021.624558}
Amit Meghanani, C.~S. Anoop, and Angarai~Ganesan Ramakrishnan.
\newblock Recognition of alzheimer’s dementia from the transcriptions of
  spontaneous speech using fasttext and cnn models.
\newblock {\em Frontiers in Computer Science}, 3:7, 2021.

\bibitem{rohanian20_interspeech}
Morteza Rohanian, Julian Hough, and Matthew Purver.
\newblock {Multi-Modal Fusion with Gating Using Audio, Lexical and Disfluency
  Features for Alzheimer’s Dementia Recognition from Spontaneous Speech}.
\newblock In {\em Proc. Interspeech 2020}, pages 2187--2191, 2020.

\bibitem{rohanian21_interspeech}
Morteza Rohanian, Julian Hough, and Matthew Purver.
\newblock {Alzheimer’s Dementia Recognition Using Acoustic, Lexical,
  Disfluency and Speech Pause Features Robust to Noisy Inputs}.
\newblock In {\em Proc. Interspeech 2021}, pages 3820--3824, 2021.

\bibitem{cummins20_interspeech}
Nicholas Cummins, Yilin Pan, Zhao Ren, Julian Fritsch, Venkata~Srikanth
  Nallanthighal, Heidi Christensen, Daniel Blackburn, Björn~W. Schuller,
  Mathew Magimai-Doss, Helmer Strik, and Aki Härmä.
\newblock {A Comparison of Acoustic and Linguistics Methodologies for
  Alzheimer’s Dementia Recognition}.
\newblock In {\em Proc. Interspeech 2020}, pages 2182--2186, 2020.

\bibitem{10.1001/archneur.1994.00540180063015}
James~T. Becker, François Boiler, Oscar~L. Lopez, Judith Saxton, and Karen~L.
  McGonigle.
\newblock {The Natural History of Alzheimer's Disease: Description of Study
  Cohort and Accuracy of Diagnosis}.
\newblock {\em Archives of Neurology}, 51(6):585--594, 06 1994.

\bibitem{macwhinney2014childes}
Brian MacWhinney.
\newblock {\em The CHILDES project: Tools for analyzing talk, Volume II: The
  database}.
\newblock Psychology Press, 2014.

\bibitem{lee-et-al-pylangacq:2016}
Jackson~L. Lee, Ross Burkholder, Gallagher~B. Flinn, and Emily~R. Coppess.
\newblock Working with chat transcripts in python.
\newblock Technical Report TR-2016-02, Department of Computer Science,
  University of Chicago, 2016.

\bibitem{devlin-etal-2019-bert}
Jacob Devlin, Ming-Wei Chang, Kenton Lee, and Kristina Toutanova.
\newblock {BERT}: Pre-training of deep bidirectional transformers for language
  understanding.
\newblock In {\em Proceedings of the 2019 Conference of the North {A}merican
  Chapter of the Association for Computational Linguistics: Human Language
  Technologies, Volume 1 (Long and Short Papers)}, pages 4171--4186,
  Minneapolis, Minnesota, June 2019. Association for Computational Linguistics.

\bibitem{lee2020biobert}
Jinhyuk Lee, Wonjin Yoon, Sungdong Kim, Donghyeon Kim, Sunkyu Kim, Chan~Ho So,
  and Jaewoo Kang.
\newblock Biobert: a pre-trained biomedical language representation model for
  biomedical text mining.
\newblock {\em Bioinformatics}, 36(4):1234--1240, 2020.

\bibitem{alsentzer-etal-2019-publicly}
Emily Alsentzer, John Murphy, William Boag, Wei-Hung Weng, Di~Jindi, Tristan
  Naumann, and Matthew McDermott.
\newblock Publicly available clinical {BERT} embeddings.
\newblock In {\em Proceedings of the 2nd Clinical Natural Language Processing
  Workshop}, pages 72--78, Minneapolis, Minnesota, USA, June 2019. Association
  for Computational Linguistics.

\bibitem{NEURIPS2020_96da2f59}
Zi-Hang Jiang, Weihao Yu, Daquan Zhou, Yunpeng Chen, Jiashi Feng, and Shuicheng
  Yan.
\newblock Convbert: Improving bert with span-based dynamic convolution.
\newblock In H.~Larochelle, M.~Ranzato, R.~Hadsell, M.~F. Balcan, and H.~Lin,
  editors, {\em Advances in Neural Information Processing Systems}, volume~33,
  pages 12837--12848. Curran Associates, Inc., 2020.

\bibitem{liu2019roberta}
Yinhan Liu, Myle Ott, Naman Goyal, Jingfei Du, Mandar Joshi, Danqi Chen, Omer
  Levy, Mike Lewis, Luke Zettlemoyer, and Veselin Stoyanov.
\newblock Roberta: A robustly optimized bert pretraining approach, 2019.

\bibitem{lan2019albert}
Zhenzhong Lan, Mingda Chen, Sebastian Goodman, Kevin Gimpel, Piyush Sharma, and
  Radu Soricut.
\newblock Albert: A lite bert for self-supervised learning of language
  representations.
\newblock {\em arXiv preprint arXiv:1909.11942}, 2019.

\bibitem{NEURIPS2019_dc6a7e65}
Zhilin Yang, Zihang Dai, Yiming Yang, Jaime Carbonell, Russ~R Salakhutdinov,
  and Quoc~V Le.
\newblock Xlnet: Generalized autoregressive pretraining for language
  understanding.
\newblock In H.~Wallach, H.~Larochelle, A.~Beygelzimer, F.~d\textquotesingle
  Alch\'{e}-Buc, E.~Fox, and R.~Garnett, editors, {\em Advances in Neural
  Information Processing Systems}, volume~32. Curran Associates, Inc., 2019.

\bibitem{NIPS2016_9dcb88e0}
Jiasen Lu, Jianwei Yang, Dhruv Batra, and Devi Parikh.
\newblock Hierarchical question-image co-attention for visual question
  answering.
\newblock In D.~Lee, M.~Sugiyama, U.~Luxburg, I.~Guyon, and R.~Garnett,
  editors, {\em Advances in Neural Information Processing Systems}, volume~29.
  Curran Associates, Inc., 2016.

\bibitem{shu2019defend}
Kai Shu, Limeng Cui, Suhang Wang, Dongwon Lee, and Huan Liu.
\newblock defend: Explainable fake news detection.
\newblock In {\em Proceedings of the 25th ACM SIGKDD international conference
  on knowledge discovery \& data mining}, pages 395--405, 2019.

\bibitem{lu-li-2020-gcan}
Yi-Ju Lu and Cheng-Te Li.
\newblock {GCAN}: Graph-aware co-attention networks for explainable fake news
  detection on social media.
\newblock In {\em Proceedings of the 58th Annual Meeting of the Association for
  Computational Linguistics}, pages 505--514, Online, July 2020. Association
  for Computational Linguistics.

\bibitem{caruana1997multitask}
Rich Caruana.
\newblock Multitask learning.
\newblock {\em Machine learning}, 28(1):41--75, 1997.

\bibitem{rajamanickam-etal-2020-joint}
Santhosh Rajamanickam, Pushkar Mishra, Helen Yannakoudakis, and Ekaterina
  Shutova.
\newblock Joint modelling of emotion and abusive language detection.
\newblock In {\em Proceedings of the 58th Annual Meeting of the Association for
  Computational Linguistics}, pages 4270--4279, Online, July 2020. Association
  for Computational Linguistics.

\bibitem{jin-aletras-2021-modeling}
Mali Jin and Nikolaos Aletras.
\newblock Modeling the severity of complaints in social media.
\newblock In {\em Proceedings of the 2021 Conference of the North American
  Chapter of the Association for Computational Linguistics: Human Language
  Technologies}, pages 2264--2274, Online, June 2021. Association for
  Computational Linguistics.

\bibitem{wolf-etal-2020-transformers}
Thomas Wolf, Lysandre Debut, Victor Sanh, Julien Chaumond, Clement Delangue,
  Anthony Moi, Pierric Cistac, Tim Rault, Rémi Louf, Morgan Funtowicz, Joe
  Davison, Sam Shleifer, Patrick von Platen, Clara Ma, Yacine Jernite, Julien
  Plu, Canwen Xu, Teven~Le Scao, Sylvain Gugger, Mariama Drame, Quentin Lhoest,
  and Alexander~M. Rush.
\newblock Transformers: State-of-the-art natural language processing.
\newblock In {\em Proceedings of the 2020 Conference on Empirical Methods in
  Natural Language Processing: System Demonstrations}, pages 38--45, Online,
  October 2020. Association for Computational Linguistics.

\bibitem{portelli-etal-2021-bert}
Beatrice Portelli, Edoardo Lenzi, Emmanuele Chersoni, Giuseppe Serra, and
  Enrico Santus.
\newblock {BERT} prescriptions to avoid unwanted headaches: A comparison of
  transformer architectures for adverse drug event detection.
\newblock In {\em Proceedings of the 16th Conference of the European Chapter of
  the Association for Computational Linguistics: Main Volume}, pages
  1740--1747, Online, April 2021. Association for Computational Linguistics.

\bibitem{benjamini1995controlling}
Yoav Benjamini and Yosef Hochberg.
\newblock Controlling the false discovery rate: a practical and powerful
  approach to multiple testing.
\newblock {\em Journal of the Royal statistical society: series B
  (Methodological)}, 57(1):289--300, 1995.

\bibitem{rissola2020beyond}
Esteban~Andr{\'e}s R{\'\i}ssola, Mohammad Aliannejadi, and Fabio Crestani.
\newblock Beyond modelling: understanding mental disorders in online social
  media.
\newblock In {\em European Conference on Information Retrieval}, pages
  296--310. Springer, 2020.

\bibitem{schwartz2013personality}
H~Andrew Schwartz, Johannes~C Eichstaedt, Margaret~L Kern, Lukasz Dziurzynski,
  Stephanie~M Ramones, Megha Agrawal, Achal Shah, Michal Kosinski, David
  Stillwell, Martin~EP Seligman, et~al.
\newblock Personality, gender, and age in the language of social media: The
  open-vocabulary approach.
\newblock {\em PloS one}, 8(9):e73791, 2013.

\bibitem{almor1999alzheimer}
Amit Almor, Daniel Kempler, Maryellen~C MacDonald, Elaine~S Andersen, and
  Lorraine~K Tyler.
\newblock Why do alzheimer patients have difficulty with pronouns? working
  memory, semantics, and reference in comprehension and production in
  alzheimer's disease.
\newblock {\em Brain and language}, 67(3):202--227, 1999.

\bibitem{watson1999analysis}
Caroline~M Watson.
\newblock An analysis of trouble and repair in the natural conversations of
  people with dementia of the alzheimer's type.
\newblock {\em Aphasiology}, 13(3):195--218, 1999.

\bibitem{garcia1997analysis}
Linda~J Garcia and Yves Joanette.
\newblock Analysis of conversational topic shifts: A multiple case study.
\newblock {\em Brain and language}, 58(1):92--114, 1997.

\bibitem{doi:10.3109/02699206.2012.751626}
Roelien Bastiaanse.
\newblock Why reference to the past is difficult for agrammatic speakers.
\newblock {\em Clinical Linguistics \& Phonetics}, 27(4):244--263, 2013.
\newblock PMID: 23339396.

\bibitem{Ribeiro_Singh_Guestrin_2018}
Marco~Tulio Ribeiro, Sameer Singh, and Carlos Guestrin.
\newblock Anchors: High-precision model-agnostic explanations.
\newblock {\em Proceedings of the AAAI Conference on Artificial Intelligence},
  32(1), Apr. 2018.

\end{thebibliography}

\end{document}